\documentclass[letterpaper]{article}
\usepackage[]{aaai23}
\nocopyright

\usepackage[numbers]{natbib}
\usepackage{cite}
\usepackage{amsmath,amssymb,amsfonts}
\usepackage{algorithmic}
\usepackage{graphicx}
\usepackage{textcomp}
\usepackage{xcolor}
\def\BibTeX{{\rm B\kern-.05em{\sc i\kern-.025em b}\kern-.08em
    T\kern-.1667em\lower.7ex\hbox{E}\kern-.125emX}}

\usepackage{afterpage}
\usepackage{adjustbox}
\usepackage{booktabs}
\usepackage{tikz}
\usepackage{tcolorbox}
\usepackage{multirow}
\usepackage{rotating}
\usepackage{array}
\usepackage{amsmath}
\usepackage{graphicx}
\usepackage{subcaption}
\usepackage{tabularx}
\usepackage{url}

\newcommand*\rot{\adjustbox{rotate=45}}
    
\begin{document}
\newcolumntype{A}{>{\scriptsize}r}

\definecolor{darkred}{RGB}{139,0,0}
\definecolor{darkgreen}{RGB}{0,139,0}

% Colors for custom figures
\definecolor{1_CLS}{RGB}{255,255,255}
\definecolor{1_venezuela}{RGB}{241,253,241}
\definecolor{1_prepares}{RGB}{250,254,250}
\definecolor{1_for}{RGB}{250,254,250}
\definecolor{1_chavez}{RGB}{210,248,210}
\definecolor{1_recall}{RGB}{253,241,241}
\definecolor{1_vote}{RGB}{134,237,134}
\definecolor{1_supporters}{RGB}{201,247,201}
\definecolor{1_and}{RGB}{250,254,250}
\definecolor{1_rivals}{RGB}{250,254,250}
\definecolor{1_warn}{RGB}{253,246,246}
\definecolor{1_of}{RGB}{255,255,255}
\definecolor{1_possible}{RGB}{246,253,246}
\definecolor{1_fraud}{RGB}{228,251,228}
\definecolor{1_semi}{RGB}{251,232,232}
\definecolor{1_government}{RGB}{237,252,237}
\definecolor{1_says}{RGB}{223,250,223}
\definecolor{1_chavez}{RGB}{223,250,223}
\definecolor{1_apos}{RGB}{241,253,241}
\definecolor{1_s}{RGB}{246,253,246}
\definecolor{1_defeat}{RGB}{223,250,223}
\definecolor{1_could}{RGB}{232,251,232}
\definecolor{1_produce}{RGB}{254,250,250}
\definecolor{1_turmoil}{RGB}{196,246,196}
\definecolor{1_in}{RGB}{237,252,237}
\definecolor{1_world}{RGB}{183,244,183}
\definecolor{1_oil}{RGB}{152,240,152}
\definecolor{1_market}{RGB}{228,251,228}
\definecolor{1_period}{RGB}{246,253,246}
\definecolor{1_SEP}{RGB}{255,255,255}

\definecolor{2_CLS}{RGB}{255,255,255}
\definecolor{2_venezuela}{RGB}{254,250,250}
\definecolor{2_prepares}{RGB}{250,254,250}
\definecolor{2_for}{RGB}{250,254,250}
\definecolor{2_chavez}{RGB}{205,247,205}
\definecolor{2_recall}{RGB}{255,255,255}
\definecolor{2_election}{RGB}{192,246,192}
\definecolor{2_supporters}{RGB}{210,248,210}
\definecolor{2_and}{RGB}{241,253,241}
\definecolor{2_rivals}{RGB}{255,255,255}
\definecolor{2_warn}{RGB}{253,246,246}
\definecolor{2_of}{RGB}{255,255,255}
\definecolor{2_possible}{RGB}{246,253,246}
\definecolor{2_fraud}{RGB}{241,253,241}
\definecolor{2_semi}{RGB}{251,228,228}
\definecolor{2_government}{RGB}{228,251,228}
\definecolor{2_says}{RGB}{219,249,219}
\definecolor{2_chavez}{RGB}{210,248,210}
\definecolor{2_apos}{RGB}{241,253,241}
\definecolor{2_s}{RGB}{246,253,246}
\definecolor{2_defeat}{RGB}{219,249,219}
\definecolor{2_could}{RGB}{228,251,228}
\definecolor{2_produce}{RGB}{255,255,255}
\definecolor{2_turmoil}{RGB}{196,246,196}
\definecolor{2_in}{RGB}{232,251,232}
\definecolor{2_world}{RGB}{165,242,165}
\definecolor{2_oil}{RGB}{125,236,125}
\definecolor{2_market}{RGB}{223,250,223}
\definecolor{2_period}{RGB}{250,254,250}
\definecolor{2_SEP}{RGB}{255,255,255}
\definecolor{2_CLS}{RGB}{255,255,255}

\definecolor{3_CLS}{RGB}{255,255,255}
\definecolor{3_venezuela}{RGB}{228,251,228}
\definecolor{3_prepares}{RGB}{255,255,255}
\definecolor{3_for}{RGB}{237,252,237}
\definecolor{3_chavez}{RGB}{196,246,196}
\definecolor{3_recall}{RGB}{219,249,219}
\definecolor{3_UNK}{RGB}{251,232,232}
\definecolor{3_supporters}{RGB}{219,249,219}
\definecolor{3_and}{RGB}{237,252,237}
\definecolor{3_rivals}{RGB}{228,251,228}
\definecolor{3_warn}{RGB}{255,255,255}
\definecolor{3_of}{RGB}{250,254,250}
\definecolor{3_possible}{RGB}{237,252,237}
\definecolor{3_fraud}{RGB}{241,253,241}
\definecolor{3_semi}{RGB}{250,223,223}
\definecolor{3_government}{RGB}{232,251,232}
\definecolor{3_says}{RGB}{228,251,228}
\definecolor{3_chavez}{RGB}{214,249,214}
\definecolor{3_apos}{RGB}{237,252,237}
\definecolor{3_s}{RGB}{241,253,241}
\definecolor{3_defeat}{RGB}{210,248,210}
\definecolor{3_could}{RGB}{232,251,232}
\definecolor{3_produce}{RGB}{254,250,250}
\definecolor{3_turmoil}{RGB}{201,247,201}
\definecolor{3_in}{RGB}{237,252,237}
\definecolor{3_world}{RGB}{170,242,170}
\definecolor{3_oil}{RGB}{130,237,130}
\definecolor{3_market}{RGB}{210,248,210}
\definecolor{3_period}{RGB}{237,252,237}
\definecolor{3_SEP}{RGB}{255,255,255}

% Reference: https://aclanthology.org/D18-1407.pdf
\definecolor{burgundy}{rgb}{0.5, 0.0, 0.13}
\definecolor{shade}{rgb}{0.3, 0.3, 0.3}
\definecolor{unkr}{HTML}{333333}
\definecolor{the}{HTML}{b8f5b8}
\definecolor{beautiful}{HTML}{20df20}
\definecolor{images}{HTML}{20df20}
\definecolor{and}{HTML}{fadbdb}
\definecolor{solemn}{HTML}{55e755}
\definecolor{words}{HTML}{f09494}
\newcommand*{\mybox}[2]{\tikz[anchor=base,baseline=0pt,rounded corners=0pt, inner sep=0.2mm] \node[fill=#1!60!white] (X) {#2};}
\newcommand*{\unkbox}[2]{\tikz[anchor=base,baseline=0pt,rounded corners=1pt, inner sep=0.2mm] \node[fill=#1!60!white] (X) {#2};}
\renewcommand*{\bibfont}{\footnotesize}
\newcommand{\UNK}{\texttt{[UNK]}}
\newcommand{\UNKR}{\unkbox{unkr}{\strut{\textcolor{white}{\texttt{[UNK]}}}}}

\newcommand{\dataset}{{\cal D}}
\newcommand{\fracpartial}[2]{\frac{\partial #1}{\partial  #2}}

\title{Robust Infidelity: When Faithfulness Measures on Masked Language Models Are Misleading}

\author {
    Evan Crothers, \textsuperscript{\rm 1}
    Herna Viktor, \textsuperscript{\rm 1}
    Nathalie Japkowicz \textsuperscript{\rm 2}
}
\affiliations {
    \textsuperscript{\rm 1} University of Ottawa, Ottawa, Ontario, Canada\\
    \textsuperscript{\rm 2} American University, Washington D.C., U.S.A.\\
    ecrot027@uottawa.ca, hviktor@uottawa.ca, japkowic@american.edu
}

\maketitle

\begin{abstract}
A common approach to quantifying neural text classifier interpretability is to calculate faithfulness metrics based on iteratively masking salient input tokens and measuring changes in the model prediction.  We propose that this property is better described as ``sensitivity to iterative masking", and highlight pitfalls in using this measure for comparing text classifier interpretability.  We show that iterative masking produces large variation in faithfulness scores between otherwise comparable Transformer encoder text classifiers. We then demonstrate that iteratively masked samples produce embeddings outside the distribution seen during training, resulting in unpredictable behaviour.  We further explore task-specific considerations that undermine principled comparison of interpretability using iterative masking, such as an underlying similarity to salience-based adversarial attacks.  Our findings give insight into how these behaviours affect neural text classifiers, and provide guidance on how sensitivity to iterative masking should be interpreted.

\end{abstract}

% \begin{keywords}
%   iterative masking, interpretability, large language models, adversarial attacks, fairness
% \end{keywords}

\section{Introduction}

Transformer models have become ubiquitous in natural language processing (NLP), achieving state-of-the-art performance across a variety of domains \citep{vaswani2017attention}. Despite this success, there remains widespread concern about the lack of transparency and explainability of neural language models \citep{10.1145/3236386.3241340}, which hinders understanding of what they learn and limits their use in high-stakes applications requiring human oversight or explanation.

Feature attributions are one method used to improve the interpretability of neural text classifiers, producing scores that indicate how much each token contributes to a given prediction \citep{10.5555/3295222.3295230, sundararajan2017axiomatic}.  These scores allow a reviewer to identify important input tokens and interpret model responses. 

A challenge arises in quantifying the quality of these interpretations.  Previous work has defined the property of ``faithfulness" as how well an explanation reflects the observed behaviour of a model, separate from human-centric considerations such as an explanation's plausibility \citep{jacovi2020towards}.  Measurement of faithfulness relies upon automated measures designed to quantify to what extent an explanation accurately reflects how a model behaves. For neural text classifiers, these measures are overwhelmingly calculated by means of iteratively removing salient features and measuring model responses \citep{atanasova-etal-2020-diagnostic, deyoung-etal-2020-eraser, nguyen2018comparing, zafar-etal-2021-lack}.

Iteratively masking features to measure faithfulness allows for comparison of feature explanations based on whether the explanation correctly ranks the tokens that cause the largest impact on the output \citep{sundararajan2017axiomatic}.  However, these measures have also been used as a means of comparing different models.  In such an approach, the same feature explanation method is applied to two models, a masking-based faithfulness measure is calculated, and the model that produces higher average scores is theorized to be more interpretable \citep{yoo2021towards}.

Our results indicate that iterative masking results in model-specific behaviours that undermine cross-model comparison.  In addition to this, we demonstrate the out-of-domain nature of iteratively masked samples by measuring the impact of masking on model embeddings.  We further determine that there are circumstances in which iteratively masking salient tokens bears a strong resemblance to a token-level adversarial attack, which raises concerns about considering sensitivity to iterative masking as an unambiguously desirable property.  

The remainder of this work is organized as follows.  Section \ref{sec:fid-rel} describes related work on evaluating faithfulness in neural text classification. Section \ref{sec:fid-exp} describes the datasets, models, and experiment settings used.  Section \ref{sec:fid-res-faith} characterizes the specific behaviours that undermine the robustness of faithfulness measures for cross-model comparison.  Section \ref{sec:fid-res-emb} shows that removing words from samples produces out-of-manifold inputs. Section \ref{sec:fid-res-adv} explores how robustness to iterative masking relates to adversarial attacks.  Section \ref{sec:fid-disc} contains discussion and recommendations based on our findings, and presents our conclusions.

\section{Related Work}
\label{sec:fid-rel}

\subsection{Feature-based Interpretability Methods}

Feature-based interpretability methods for deep learning models, such as LIME \citep{ribeiro2016should}, SHAP \citep{10.5555/3295222.3295230}, and integrated gradients \citep{sundararajan2017axiomatic} assign an importance score to input features to determine their contribution to a particular network result.  Evaluation of these interpretability methods has shown that gradient-based approaches demonstrate the best agreement with human assessment for Transformer models, as well as best correlating with tokens which cause the greatest drop in performance if they are removed from the model \citep{atanasova-etal-2020-diagnostic}.

Based on these findings, we use integrated gradients to generate feature attributions in our work.  Integrated gradients is a strong axiomatic method for calculating how input features contribute to the output of a model \citep{sundararajan2017axiomatic}.  This method interpolates between a baseline input representation $x'$ (in this case a zero embedding vector) and the embedding vectors $x$ of each token.

The robustness of feature-based interpretabilty methods for neural text classifiers has been questioned \citep{madsen2022evaluating, zafar-etal-2021-lack}.  Specifically, it has been demonstrated that 1) two functionally near-equivalent models with differing weight initializations may produce different explanations, and 2) feature attributions of a model with random parameters may be the same as for a model with learned parameters.  

In our work, we demonstrate the mechanisms by which breakdowns in faithfulness measures occur, and demonstrate that while measures based on iterative masking may be useful for characterizing model responses in certain situations, such as under adversarial attacks, they are not generally appropriate for cross-model comparison of interpretability.

\subsection{Faithfulness Measures}

Recall that a key method for evaluating the quality of interpretable explanations are faithfulness measures.  Faithfulness measures are commonly calculated by iteratively hiding features in descending order of feature importance and determining a score based on changes in model output.  These scores may be calculated by removing features until the output classification changes \citep{nguyen2018comparing, zafar-etal-2021-lack} or by removing a preset number of salient tokens and comparing the change in class probabilities \citep{deyoung-etal-2020-eraser, atanasova-etal-2020-diagnostic}.  

\subsubsection{Fidelity Calculation}

Our approach to quantifying faithfulness closely aligns with \citep{arras-etal-2016-explaining} and \citep{zafar-etal-2021-lack}, in which tokens are masked from the input in descending order of feature attribution scores until the result of the model changes.  We then record the \% of tokens that were removed at the point that the model's output changes.

Formally, given an input text split into $n$ tokens, $T = [t_1, ..., t_n]$, a vector of feature explanations $\Phi(T) = [\phi(t_1), ..., \phi(t_n)]$, a model $m$, and the model's unknown vocabulary token $\texttt{[UNK]}$, we first calculate $m(T) = y_0$.  We then define an iterative scoring function $f(T)$, that at each step performs the replacement $T[\texttt{max}\ \Phi(t)] \leftarrow \texttt{[UNK]} = T'$, and calculates $m(T') = y'$.  We iterate $C$ times until $y' \neq y_0$, and return the ratio $f(T)=\frac{C}{N}$.  $f(T)$ is calculated for all $K$ input texts, and we calculate the fidelity score for the model as:

\begin{equation}
Fidelity(m) = 1 - \frac{1}{K}\sum_{k=1}^{k} f(T_k)
\end{equation}

We rely on fidelity due to its simplicity and the lack of \textit{a priori} parameters.  For completeness, we now briefly describe the calculation of measures that rely on masking set numbers of tokens so that we can refer to them in explaining theoretical pitfalls in cross-model comparison.

\subsubsection{Area over the perturbation curve}
A common faithfulness measure that relies on masking a preset number of tokens is the ``area over the perturbation curve" (AOPC) \citep{samek2017, nguyen2018comparing}. AOPC involves first creating an ordered ranking of input tokens by importance $x_1, x_2, ... x_n$ in each $n$-token input sequence. AOPC is then calculated by:

\begin{equation}
AOPC = \frac{1}{L+1} \left \langle \sum_{k=1}^{L} f(x) - f(x_{1..k}) \right \rangle_{p(x)}
\label{eq:aopc}
\end{equation}

where $L$ is the \textit{a priori} number of tokens to mask, $f(x_{1..k})$ is the output probability for the original predicted class when tokens $1..k$ are removed, and $\langle \cdot \rangle_{p(x)}$ denotes the average over all sequences in the dataset.  We refer to this calculation when discussing theoretical weaknesses of cross-model faithfulness comparison based on iterative masking.

\subsubsection{Limitations of Faithfulness Measures}

While faithfulness measures can be used to compare different methods of generating explanations \citep{nguyen2018comparing,atanasova-etal-2020-diagnostic}, recent research suggests that such measures may not be suitable for cross-model comparison of interpretability.

Specifically, recent work demonstrates that untrained models produce fidelity scores well above random masking, and the fidelity of the same interpretability method applied across different encoders can vary substantially \citep{zafar-etal-2021-lack}.  This previous analysis left investigation of the root cause of these phenomena as future work.  We shed light on this root cause by demonstrating that the iterative masking process produces samples outside the manifold of the training data, resulting in differing model-specific behaviour.

Producing reliable explanations in the text domain has been highlighted as difficult due to aberrant behaviour on iteratively masked samples \citep{feng-etal-2018-pathologies}.  By removing the least important word from a sequence iteratively, the resulting condensed explanation is no longer meaningful to human observers, impacting human-assessed plausibility. Our work complements this, highlighting how iterative masking affects representations within neural text classifiers and impacts automated assessment of faithfulness.

\subsection{Adversarial Training and Faithfulness}

%Adversarial attacks --- inputs perturbed to cause a model to produce an erroneous misclassification --- can be applied in the text domain \citep{jin2020bert}. A common word-based approach is to replace words with synonyms, typically by using a synonym dictionary, or by leveraging another language model to find nearby words with nearby embeddings \citep{alzantot-etal-2018-generating, shi2019robustness}.  The attacks DeepWordBug \citep{gao2018black} and HotFlip \citep{ebrahimi-etal-2018-hotflip} introduce targeted character-level perturbations to cause erroneous classifications.

%It has been shown that input feature attributions can be used as an effective feature for detection of adversarial attacks in the text domain \citep{huber2022detecting}.

Previous work has investigated the impact of training on adversarial samples on interpretability measures, and found that adversarial training appears to increase the faithfulness measures of models \citep{yoo2021towards} and alignment with ground truth explanations \citep{sadria2023adversarial}. In the process of comparing faithfulness of neural text classifiers after adversarial training, one previous work found that RoBERTa models scored significantly lower than BERT models, and theorized that RoBERTa may be less interpretable than BERT \citep{yoo2021towards}. 

We demonstrate in our research that increased faithfulness measures are not universal in the presence of adversarial attacks.  Further, our overall findings suggest that previously observed differences between BERT and RoBERTa faithfulness scores are not a meaningful reflection of each models inherent interpretability, but rather a result of model-specific behaviours in the presence of iterative masking. 
\section{Datasets and Experimental Setup}
\label{sec:fid-exp}

To ensure a reproducible set of task-specific models and associated data samples, we use the text classification attack benchmark (TCAB) dataset and models \citep{asthana2022tcab}.  This benchmark includes: 1) a number of established NLP task datasets; 2) BERT and RoBERTa models trained for each task; and 3) a variety of successful adversarial attacks against the included models.  

To obtain a range of data domains and sequence lengths, we perform our experiments on the following task datasets in TCAB: 1) the Stanford Sentiment Treebank (SST-2) dataset of movie reviews for sentiment classification, a common NLP benchmark task \citep{socher2013recursive}; 2) the Twitter climate change sentiment dataset, a multi-class dataset of social media data \citep{qian_2019}; 3) Wikipedia Toxic Comments, a dataset with a longer sequence length widely-used for studying adversarial attacks \citep{dixon2018measuring}; and 4) the Civil Comments dataset, a dataset of online comments for evaluating unintended bias in toxicity detection, an area where interpretability may be important \citep{jigsaw_2019}.

The Transformer language models included in this benchmark reflect common architectures for contemporary neural text classifiers.  We consider both a baseline bi-directional Transformer architecture BERT \citep{devlin-etal-2019-bert}, as well as the commonly used robustly-optimized variant of this architecture, RoBERTa \citep{Liu2019RoBERTaAR}.

In Section \ref{sec:fid-res-adv}, we select a variety of adversarial attacks to analyze, including well-known word-level and character-level attacks to compare any differences in the impacts of adversarial training on fidelity.  Specifically, we use DeepWordBug \citep{gao2018black}, TextFooler \citep{jin2020bert}, Genetic \citep{alzantot-etal-2018-generating}, and HotFlip \citep{ebrahimi-etal-2018-hotflip} attacks.

As calculating input attributions is computationally expensive, fidelity calculations in Table \ref{tab:fidelityratios} are based on a sample of 1,600 total records evenly split across dataset-model combinations.  Initial experiments indicated this was sufficient to observe consistent patterns in fidelity scores, while freeing computational resources to explore multiple combinations.  We use $N=30$ for the number of steps for layer integrated gradient calculation, as this approximates the largest value that fits within the memory constraints of the system, and higher step counts typically produce more accurate explanations \citep{sundararajan2017axiomatic}. To evaluate models in the presence of class imbalance, we use the macro F1 score.

All experiments were performed on a Windows workstation with an Intel i7-6800K 12-core CPU, 64GB RAM, and a 24GB VRAM NVIDIA GPU.  Fixed seeds are used throughout experiments for reproducibility. Complete experiment code, along with references to utilized models and data are provided to ensure reproducible results.

\section{Pitfalls in Cross-Model Comparison}
\label{sec:fid-res-faith}

Faithfulness measures based on iterative masking produces scores that are sensitive to model initialization, with identical models of the same architecture and comparable dataset performance producing dramatically differing scores \citep{zafar-etal-2021-lack}. We explore the mechanisms that result in these pitfalls by first explaining the theoretical deficiencies in the iterative masking approach, illustrated with an example, and then demonstrate the variation in behaviour from 8 models trained across 4 task datasets. In doing so, we explore the mechanisms that enable faithfulness metrics to exhibit substantial variation between comparable models, and their high sensitivity to characteristics of the training dataset.

Figure \ref{fig:masking} demonstrates an excerpt from a positive movie review, which when iteratively masked, at no point causes the sentiment classifier to determine that the movie review is negative.  Integrated gradients feature attributions indicate which input tokens were most important to the output classification, ranking the tokens ``beautiful", ``images", and ``solemn" as most important to the model's output.  After masking these three words, the model's class confidence is at a minimum, but the output class is unchanged.

\begin{figure}[h]
\begin{adjustbox}{center}
\scriptsize
\tikz\node[fill=white!90!black,inner sep=1pt,rounded corners=0.2cm]{
\begin{tabular}{p{.08\columnwidth}p{.08\columnwidth}p{.75\columnwidth}}
Neg & Pos & Sample under iterative masking; iterative deletion \\\midrule
0.09 & \textbf{99.1} & \mybox{the}{\strut{The}} \mybox{beautiful}{\strut{beautiful}} \mybox{images}{\strut{images}} \mybox{and}{\strut{and}} \mybox{solemn}{\strut{solemn}} \mybox{words}{\strut{words}} \\
\midrule
0.24 & \textbf{97.6} & The \UNKR{} images and solemn words\\

3.6 & \textbf{96.4} & The \UNKR{} \UNKR{} and solemn words\\

28.3 & \textbf{71.7} & The \UNKR{} \UNKR{} and \UNKR{} words\\

14.0 & \textbf{86.0} & \UNKR{} \UNKR{} \UNKR{} and \UNKR{} words\\

5.2 & \textbf{94.8} & \UNKR{} \UNKR{} \UNKR{} \UNKR{} \UNKR{} words\\  \midrule

%2.3 & \textbf{97.8} & \UNKR{} \UNKR{} \UNKR{} \UNKR{} \UNKR{} \UNKR{}\\ \midrule

 1.4 & \textbf{98.6} & The images and solemn words\\

 3.6 & \textbf{96.4} & The and solemn words \\

 12.0 & \textbf{88.0} & The and words \\

 5.8 & \textbf{94.2} & and words \\

 5.5 & \textbf{94.6} & words \\

% This final result actually adds to the effect -- you can't predict that it's going to go one way or another monotonically.
% Either way though, it's a worst-case fidelity score.  Flipping your classification at 100% masking is still a score of 0.
%\textbf{69.2} & 30.8 & \textit{$<$empty$>$}
%\textbf{69.2} & 30.8 & \textit{$<$\texttt{[CLS]},  \texttt{[SEP]} $>$}
\end{tabular}
};
\end{adjustbox}
\caption{Iterative token removal in descending order of feature importance on a sample from SST-2.  Despite identifying the most important tokens, the classification is unchanged during either iterative masking or iterative deletion.} 
\label{fig:masking}
\end{figure}

The point at which a classifier changes its predicted class (if at all), may vary substantially between models.  Similarly, while removing the top $K$ tokens from the input, samples become increasingly perturbed, and output probabilities may skew back towards the original class, as is the case in Figure \ref{fig:masking}.  Area-based faithfulness measures such as AOPC are then similarly impacted as output probabilities on iteratively masked samples are not consistent across models, leading to variation in the $f(x)-f(x_{1..k})$ term (see Equation \ref{eq:aopc}). 

Under faithfulness measures that mask tokens until a change in predicted class, samples that do not cause a change in predicted class during masking incur maximum penalty to the faithfulness score, regardless of the quality of the explanation.  For methods that mask an \textit{a priori} set number of tokens, model-specific output calibration similarly undermines comparison between models \citep{guo2017calibration}.  Output logits on out-of-domain samples that include small numbers of tokens or empty strings are difficult to predict, and even if evaluating using a simple balanced binary classification dataset, there is no guarantee that this behavior will be consistent or symmetric when masking explanations for one predicted class versus the other.

Our illustrative example, in which the classification of a sample does not change during iterative masking, reflects a common situation.  Table \ref{tab:origperf} shows the performance of 8 classifiers on clean (without adversarial perturbation) samples from the TCAB dataset.  These models largely have comparable performance, with an edge to RoBERTa on the climate-change dataset.  However, substantial variation is observed across fidelity measures, highlighting the impact of both the model and the task dataset.  

To understand why discrepancies arise, we consider the failure case that we demonstrated in Figure \ref{fig:masking} --- situations where the input class never changes.  Table \ref{tab:origperf} also shows the frequency of samples where iterative masking did not lead to a change in classification result at any point. 

\begin{table}[ht]
\scriptsize
\centering
\begin{adjustbox}{center}
%\begin{tabularx}{0.8\textwidth}{lrrrr}
\begin{tabular}{lrrrr}
%\begin{tabularx}{\textwidth}{@{}l*{8}{>{\centering\arraybackslash}X}@{}}
\toprule
{} &    SST-2 & WikiTox & Civil & Climate \\
\midrule
\textbf{Macro F1: BERT }       &  91.0 &     90.3 &          83.6 &          65.0 \\
\textbf{Macro F1: RoBERTa }     &    92.0 &     90.7 &          83.0 &          74.6 \\ \midrule
\textbf{Fidelity: BERT } &  49.1 &     30.4 &          10.3 &          67.8 \\ 
\textbf{Fidelity: RoBERTa } & 45.5 &     11.8 &          5.0 &          72.2 \\\midrule
\textbf{Non-Pert Freq: BERT }  &  35.0 &     9.0 &          88.0 &          5.5  \\
\textbf{Non-Pert Freq: RoBERTa }  &  35.5 &     86.0 &          94.0 &          14.0 \\
\bottomrule
\end{tabular}
\end{adjustbox}
\caption{Behaviour of 8 task-specific TCAB BERT and RoBERTa models on unperturbed validation set, including macro F1 scores, fidelity, and frequency of samples that did not change prediction during iterative masking.} \label{tab:origperf}
\end{table}

The largest gap between models in both fidelity and frequency of samples where the predicted class was unchanged was observed on the Wikipedia Toxic Comments dataset.  The TCAB BERT and RoBERTa models trained on different datasets have comparable performance on the original classification problem, shown in Table \ref{tab:origperf}, yet the fidelity scores differ substantially. The non-perturbation frequency results provide deeper insight: the RoBERTa-wikipedia model did not change its result at any point during iterative masking on 86\% of samples.

Fundamentally, the assumption that removing salient tokens should cause the output of a model to change is not intuitive for all datasets. The Wikipedia Toxic Comments dataset is an illustrative example of this.  The class distribution of TCAB validation set for the two-class wikipedia dataset is 90.84\% negative samples, i.e., mostly comments that are not considered toxic.  Removing salient tokens one by one from an inoffensive comment is highly unlikely to create a true toxic sample, no matter how many tokens are removed.

As a result, a low fidelity score on this dataset arguably demonstrates a \textit{beneficial} property of the RoBERTa model --- resilience against adversarial attack.  The attack in this case being the removal of targeted salient tokens with the goal of changing the predicted class with the minimum number of removed tokens (even though the removal does not affect the true class).  Whether removing salient tokens is expected to affect the true class is entirely dependent on the task and dataset --- undermining the idea that iterative masking should be framed consistently across models.

We conclude that it is important to consider the dataset when assessing the interpretability of a model using a measure that penalizes robustness against perturbations.  When taken together with previous research that shows substantial variation based on model initialization \citep{zafar-etal-2021-lack}, it becomes apparent that while faithfulness measures are meaningful for evaluating the quality of explanations on the same model, they are not appropriate for cross-model comparisons of interpretability in neural text classifiers.

\section{Embeddings of Masked Samples}
\label{sec:fid-res-emb}

\begin{figure*}[h]
  \centering
  \includegraphics[width = 0.99\linewidth]{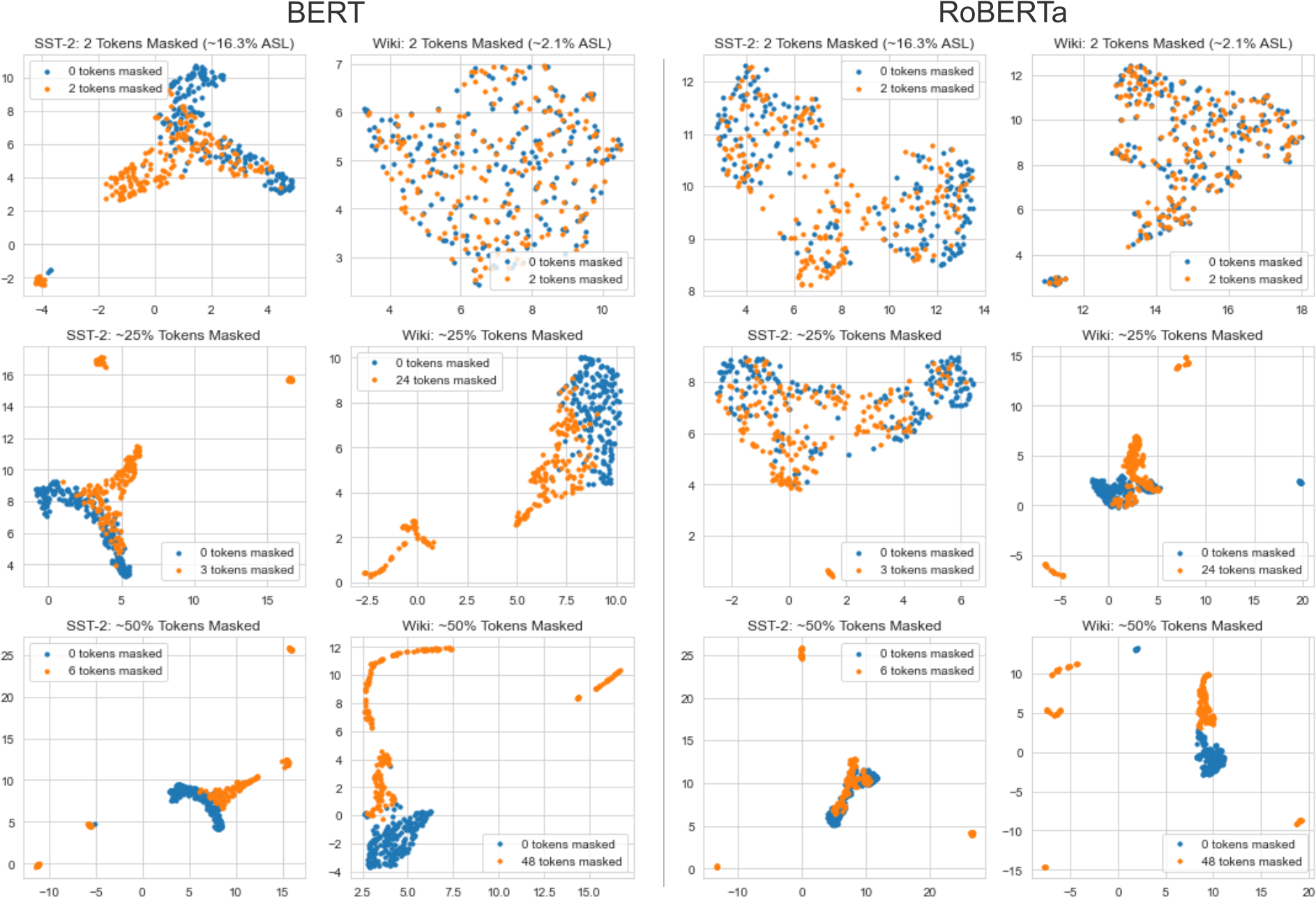}
  \vspace{0.01mm}
  \caption{UMAP projections of sample embeddings at varying levels of masking. Masking more tokens moves the resulting embeddings further out of domain of the original dataset. Masking a couple tokens within a dataset with a longer average sequence length has a relatively minor effect (e.g., see the Wikipedia Toxic Comments examples), but longer samples still generally require a significant portion of tokens to be masked to change classification (see Table \ref{tab:fidelityratios})}
  \label{fig:mask_diagram}
\end{figure*}

With the underlying mechanisms that undermine iterative masking for model comparison established, we now show that iteratively masked samples frequently fall outside the training distribution of the dataset, particularly at the levels of masking required for changes in predicted class.  As samples from this distribution haven't been seen during training, this likely leads to the inconsistent behaviour on masked samples observed in the previous section.

\subsubsection{Distributional Characteristics}

Masking tokens causes the perturbed input samples to have representations that are very different from ordinary real samples in their training datasets. To demonstrate this, we generate embeddings for each input sequence across all samples taken from each dataset by mean-pooling the output of the second-last layer of a Transformer encoder, a common approach that generally outperforms using the final-layer \texttt{[CLS]} token embeddings \citep{xiao2018bertservice, devlin-etal-2019-bert}.

Using this embedding method, we create embedding vectors $V_d = {v_1, v_2, \ldots, v_k}$ using the BERT and RoBERTa models for every sample across each dataset $d$ within our sample.  Each vector can be represented as $v_i = [x_{i1}, x_{i2}, \ldots, x_{in}]$.  We then calculate centroids $\mu_d$ for each dataset

\begin{equation}
\mu_d = \left[ \frac{\sum x_{i1}}{k}, \frac{\sum x_{i2}}{k}, \ldots, \frac{\sum x_{in}}{k} \right]
\end{equation}

where \(\sum x_{ij}\) represents the sum of the \(j\)-th component of all vectors in the set $V_d$, and the model's encoder layer dimension $n=768$ is the length of the embedding vector.  Iteratively masking tokens from each sample, we obtain an embedding vector $\omega$ at each step, and calculate cosine similarity as a scale-invariant vector similarity measure \citep{reimers2019sentence}:

\begin{equation}
    cos\_sim = \frac{{\sum_{i=1}^{n} (\omega[i] \cdot \mu[i])}}{{\sqrt{\sum_{i=1}^{n} (\omega[i])^2} \cdot \sqrt{\sum_{i=1}^{n} (\mu[i])^2}}}
\end{equation}

\begin{figure}[ht]
\centering
\includegraphics[width=0.499\textwidth]{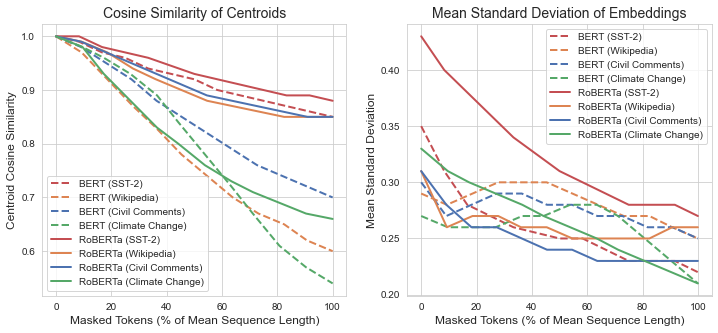}
\caption{Comparison of centroid cosine similarity and mean standard deviation of embedding vectors between BERT and RoBERTa across TCAB datasets. }
\label{fig:comparison_chart}
\end{figure}

We analyze the cosine similarity of centroids and the mean standard deviation of embedding vectors as the number of masked tokens increases, presenting the results in Figure \ref{fig:comparison_chart}, and comparing changes between models in Table \ref{tab:deltacentroid}.

\begin{table}[]
\scriptsize
\begin{adjustbox}{center}
\begin{tabular}{@{}lllllllll@{}}
\toprule
        & \multicolumn{2}{c}{SST-2}  & \multicolumn{2}{c}{WikiToxic}  & \multicolumn{2}{c}{Civ. Com.}  & \multicolumn{2}{c}{Clim. Cha.}                                                     \\ \midrule
        & $\Delta\mu$ & $\Delta\overline{\sigma}$ & $\Delta\mu$ & $\Delta\overline{\sigma}$ & $\Delta\mu$ & $\Delta\overline{\sigma}$ & $\Delta\mu$ & $\Delta\overline{\sigma}$ \\  \cmidrule(lr){2-3} \cmidrule(lr){4-5} \cmidrule(lr){6-7} \cmidrule(lr){8-9}
\textbf{BERT}    &  0.15  &  -0.13 & 0.40 & -0.04 & 0.30  & -0.05 & 0.46  & -0.06 \\
\textbf{RoBERTa} &  0.12  & -0.16 & 0.15 & -0.05 & 0.15 & -0.08 & 0.34 & -0.12                                             \\ \bottomrule
\end{tabular}
\end{adjustbox}
\caption{Change in centroid $\Delta\mu$ and mean feature standard deviation $\Delta\overline{\sigma}$ at ~50\% mean sequence length tokens masked.}
\label{tab:deltacentroid}
\end{table}

From Table \ref{tab:deltacentroid} and Figure \ref{fig:comparison_chart}, we can see that in all cases there is an increased cosine distance between the original dataset as tokens are masked.  As more tokens are replaced with \texttt{[UNK]}, we see a smaller mean standard deviation across embedding features $\Delta\overline{\sigma}$, implying that the increased presence of a single token is leading to more homogenous representations overall.  Importantly, we note substantial model-specific differences in the centroids of embeddings between unmasked and masked samples for BERT and RoBERTa.  On average, the internal representations of Wikipedia Toxic Comments samples changed less during masking for RoBERTa than for BERT, which may explain the high frequency of samples that did not change predicted class previously observed in Table \ref{tab:origperf}.

%The left plot shows centroid cosine similarity, demonstrating the shift of data representations as tokens are masked. The right plot shows the mean standard deviation of the embeddings, showing representations of partially-masked inputs are less varied.

\subsubsection{Local and Global Structure}
In addition to distributional characteristics, we can demonstrate the difference in local and global structure between unmasked and partially-masked sample embeddings using UMAP dimensionality reduction \citep{McInnes2018UMAPUM}.  We use the BERT and RoBERTa embeddings before and after masking some number of tokens as the base for these visualizations.  We choose two illustrative datasets to visualize against: SST-2 and Wikipedia Toxic Comments.  We select these two to show how the length of the input sample impacts per-token sensitivity to masking.  Our sample from the former dataset has an average sequence length of 12.25 tokens, while the latter has an average sequence length of 97.22 tokens.  Results from our dimensionality reduction can be seen in Figure \ref{fig:mask_diagram}.

We show that on datasets with short input samples such as SST-2, even just masking two salient tokens creates representations far outside the data manifold on which the model was trained, implying undefined behaviour.  The per-token impact of masking is lesser on samples with longer average sequence lengths, as shown by the Wikipedia Toxic Comments visualizations, but these longer samples also require a larger number of tokens to be masked to change the model classification.  For example, our results in Table \ref{tab:origperf} give a fidelity score of 0.304 for BERT on the Wikipedia dataset.  This indicates that on average it is required to mask 69.6\% of tokens from a Wikipedia sample before the output classification of the model changes. For RoBERTa, this proportion is even larger --- a fidelity score of 0.118 on the dataset implies that 88.2\% of tokens must be masked on average to perturb the classification result.

Based on the magnitude of the deviation, we infer that the departure of masked samples from the data manifold of the training set may be responsible for any prediction ``crossover point", particularly for datasets where the true class is likely to be unchanged by masking (such as when masking non-toxic samples in Wikipedia Toxic Comments).  Calculating faithfulness metrics using this approach thus relies heavily on undefined model-specific behaviour on degenerate samples from an unseen manifold --- behaviour which is difficult to predict, and may vary dramatically between different models.

\section{Fidelity Under Adversarial Attack}
\label{sec:fid-res-adv}

%Adversarial training, the process of training a model on adversarial inputs to improve robustness against attacks, is a major area of research in NLP \citep{bai2021recent}.
Recall that previous work has shown some small-scale experiments that suggest that adversarial training appears to improve faithfulness measures in neural text classifiers \citep{yoo2021towards}.  We perform a large-scale analysis that suggests that differences in fidelity after adversarial training or between different models do not follow easily discernible patterns, raising questions as to whether this pattern holds generally for neural text classifiers, or only under certain conditions.

For adversarial training, we perform hyperparameter tuning to determine appropriate combinations of learning rate $lr\in[10^{-6},10^{-3}]$, weight decay $wd \in [10^{-5}, 10^{-2}]$, and training epochs $ep \in [0,5]$.  Tuning of hyperparameters is required due to differences across task datasets and model architectures.  Evaluation for hyperparameter tuning was performed by performing a 90-10 train-validation split of the TCAB adversarial attack train dataset, with each training set composed of half original task samples and half adversarial samples. A batch size of 16 was set for efficient training within GPU memory limits. %We present findings based on the hyperparameter combination resulting in the best F1 score on the validation dataset.  Appendix Tables \ref{tab:f1advfaithpre} and \ref{tab:f1advfaithpost} show the F1 scores before and after adversarial training respectively, confirming the training was successful at improving adversarial robustness.

\begin{table*}[ht]
\footnotesize
\centering
\begin{adjustbox}{center}
\begin{tabular}{Arcccccccc}
\toprule
& {} & \multicolumn{4}{c}{\textbf{BERT}} & \multicolumn{4}{c}{\textbf{RoBERTa}} \\
& {} &    SST-2 & WikiTox & Civil & Climate &     SST-2 & WikiTox & Civil & Climate \\\cmidrule(lr){3-6} \cmidrule(lr){7-10}
& Clean       &  49.1 &     30.4 &          10.3 &          67.8 &   45.5 &     11.8 &          5.0 &          72.2 \\\cmidrule(lr){3-6} \cmidrule(lr){7-10}
\multirow{4}{*}[5.5ex]{\begin{tabular}[c]{@{}r@{}}{\rot{\tiny \parbox{2.5cm}{4.1) Adv. Samples\\Pre Adv. Training}}}\end{tabular}} & DeepWordBug &  69.0 &     70.3 &          18.8 &          81.8 &   68.0 &     42.3 &          13.5 &          55.2 \\
& TextFooler  &  87.9 &     74.1 &          34.1 &          80.6 &   77.4 &     53.3 &          19.2 &          73.1 \\
& Genetic     &  79.8 &     78.9 &          31.8 &          87.1 &   56.7 &     51.1 &          20.1 &          51.9 \\
& HotFlip     &  77.1 &     79.3 &          47.3 &          86.6 &   57.6 &     53.5 &          20.5 &          47.9 \\\cmidrule(lr){3-6}\cmidrule(lr){7-10} 
\multirow{4}{*}[5.5ex]{\begin{tabular}[c]{@{}r@{}}{\rot{\tiny \parbox{2.5cm}{4.2) Adv. Samples\\Post Adv. Training}}}\end{tabular}} & DeepWordBug &  59.7 &     36.2 &          8.9 &          79.7 &   65.2 &     48.3 &          25.3 &          79.4 \\
& TextFooler  &  77.5 &     47.5 &          80.7 &          70.8 &   74.2 &     52.0 &          20.1 &          77.2 \\
& Genetic     &  66.2 &     86.0 &          75.2 &          83.4 &   54.5 &     57.9 &          40.0 &          86.5 \\
& HotFlip     &  69.5 &     60.8 &          59.9 &          79.6 &   39.9 &     24.1 &          1.0 &          79.8 \\\cmidrule(lr){3-6}\cmidrule(lr){7-10} 
\multirow{4}{*}[5.5ex]{\begin{tabular}[c]{@{}r@{}}{\rot{\tiny \parbox{2.5cm}{4.3) Clean Samples\\Post Adv. Training}}}\end{tabular}} & DeepWordBug &  49.9 &     64.3 &          80.5 &          65.8 &   49.2 &     51.4 &          84.6 &          72.4 \\
& TextFooler  &  45.4 &     62.8 &          47.6 &          69.4 &   47.6 &     47.4 &          77.0 &          68.8 \\
& Genetic     &  46.4 &     41.2 &          41.7 &          66.3 &   42.0 &     28.0 &          60.3 &          72.3 \\
& HotFlip     &  47.9 &     62.4 &          32.1 &          66.7 &   47.7 &     61.3 &          86.9 &          69.7 \\
\bottomrule
\end{tabular}
\end{adjustbox}
\caption{Fidelity scores of TCAB BERT and RoBERTa classifiers under varying adversarial attacks, and fidelity of 32 adversarially trained models for each dataset-model-attack combination on adversarial and clean samples.} \label{tab:fidelityratios}
\end{table*}

In Table \ref{tab:fidelityratios}, we show fidelity calculations on 1) successful adversarial attacks prior to adversarial training; 2) adversarial samples after adversarial training; and 3) clean (non-adversarial) samples after adversarial training.  We also include fidelity scores on clean samples prior to adversarial training, previously reported in Table \ref{tab:origperf}, at the top of Table \ref{tab:fidelityratios} for ease of comparison.  %The F1 scores on TCAB test sets before and after training can be found in the Appendix.

From the results in Table \ref{tab:fidelityratios}.1, we can see that fidelity of explanations is generally higher on successful adversarial samples compared to clean samples.  Adversarial attacks are typically optimized to minimize the number of perturbations, while still altering model output.  These constraints lend themselves towards perturbed sequences where a small portion of tokens have a significant influence on predictions.  Without an attack approach that also attacks model interpretability methods \citep{ivankay2022fooling}, these highly salient perturbed tokens are identified by the feature attributions, and masked early during fidelity calculation.  As such, the model output often returns to the original value more quickly, leading to higher fidelity scores overall.

The only model where adversarial attack samples did not universally manifest in increased fidelity scores, appears to be the RoBERTa climate-change model.  In this case, the fidelity score on clean samples was already the highest of all included models, indicating that this model already relied on a relatively small number of salient tokens to make correct predictions.  Adversarial attacks here may produce perturbations that interfere with these salient tokens, resulting in attributions spread more evenly across the remaining tokens.

We further note from Table \ref{tab:fidelityratios}.1 that fidelity scores of BERT models under successful adversarial attacks prior to adversarial training are generally higher than those calculated on RoBERTa.  We view this result not as a proxy measure of interpretability, but instead as a measure of sensitivity to iterative masking --- the behaviour which the metric directly measures.  As we are working with a dataset of successful adversarial attack samples, the difference in fidelity score indicates that BERT models return to the original class after masking fewer salient tokens than RoBERTa models. That is, successful attacks on the TCAB BERT models depend on a comparatively smaller number of salient tokens. In framing fidelity this way, we demonstrate how faithfulness measures might be used to better understand how adversarial attacks impact salient tokens in neural text classifiers, rather than used as a proxy measure for explainability.

Iterative masking itself resembles a simple token-level adversarial attack.  Instead of replacing a salient token with an equivalent that causes a change in predicted class, salient tokens are masked until a change in predicted class occurs.  As long as token removal does not influence the true class, this meets the definition of an adversarial attack \citep{jin2020bert}. Such an attack may noticeably degrade the original sentence, though reduction in text quality has been observed for other word-level and character-level attacks as well \citep{crothers2022adversarial}.  Figure \ref{fig:comparison_adv_mask} demonstrates an AGNews sample where both adversarial attack and saliency-based masking target the same token (``vote") and similarly alter the predicted class.

Taken together, Tables \ref{tab:fidelityratios}.2 and \ref{tab:fidelityratios}.3, which show fidelity after adversarial training on adversarial samples and clean samples respectively, demonstrate that adversarial training of neural text classifiers does not appear to have a consistent impact on fidelity scores across different datasets.  The observed fidelity gaps can be very large, even for the same encoder, such as the BERT results for the Civil Comments dataset.  From this, we conclude that training neural text classifiers on adversarial samples does not have a straightforward relationship with sensitivity to iterative masking, despite previous indications to the contrary.

\begin{figure}[h]
\scriptsize
\begin{adjustbox}{center}
\tikz\node[draw=gray,fill=white!100!black,inner sep=2pt,rounded corners=0.2cm]{
\begin{tabular}{p{.09\columnwidth}p{.12\columnwidth}p{.72\columnwidth}}
Sample & Predicted Class &  Token Importance (True Class Attribution) \\\midrule
Clean & \textcolor{darkgreen}{World News} & \mybox{1_CLS}{[CLS]} \mybox{1_venezuela}{venezuela} \mybox{1_prepares}{prepares} \mybox{1_for}{for} \mybox{1_chavez}{chavez} \mybox{1_recall}{recall}  \mybox{1_vote}{\boxed{\textbf{vote}}}  \mybox{1_supporters}{supporters}  \mybox{1_and}{and}  \mybox{1_rivals}{rivals}  \mybox{1_warn}{warn}  \mybox{1_of}{of}  \mybox{1_possible}{possible}  \mybox{1_fraud}{fraud}  \mybox{1_semi}{;} \mybox{1_government}{government}  \mybox{1_says}{says}  \mybox{1_chavez}{chavez}  \mybox{1_apos}{'} \mybox{1_s}{s} \mybox{1_defeat}{defeat}  \mybox{1_could}{could}  \mybox{1_produce}{produce}  \mybox{1_turmoil}{turmoil}  \mybox{1_in}{in}  \mybox{1_world}{world}  \mybox{1_oil}{oil}  \mybox{1_market}{market}  \mybox{1_period}{.} \mybox{1_SEP}{[SEP]}\\\addlinespace\midrule\addlinespace
Attack & \textcolor{darkred}{Business News} & \mybox{2_CLS}{[CLS]} \mybox{2_venezuela}{venezuela} \mybox{2_prepares}{prepares} \mybox{2_for}{for} \mybox{2_chavez}{chavez} \mybox{2_recall}{recall}  \mybox{2_election}{\boxed{\textbf{election}}}  \mybox{2_supporters}{supporters}  \mybox{2_and}{and}  \mybox{2_rivals}{rivals}  \mybox{2_warn}{warn}  \mybox{2_of}{of}  \mybox{2_possible}{possible}  \mybox{2_fraud}{fraud}  \mybox{2_semi}{;} \mybox{2_government}{government}  \mybox{2_says}{says}  \mybox{2_chavez}{chavez}  \mybox{2_apos}{'} \mybox{2_s}{s} \mybox{2_defeat}{defeat}  \mybox{2_could}{could}  \mybox{2_produce}{produce}  \mybox{2_turmoil}{turmoil}  \mybox{2_in}{in}  \mybox{2_world}{world}  \mybox{2_oil}{oil}  \mybox{2_market}{market}  \mybox{2_period}{.} \mybox{2_SEP}{[SEP]}\\\midrule\addlinespace
Masked & \textcolor{darkred}{Business News} & \mybox{3_CLS}{[CLS]} \mybox{3_venezuela}{venezuela} \mybox{3_prepares}{prepares} \mybox{3_for}{for} \mybox{3_chavez}{chavez} \mybox{3_recall}{recall}  \mybox{3_UNK}{\boxed{\textbf{[UNK]}}}  \mybox{3_supporters}{supporters}  \mybox{3_and}{and}  \mybox{3_rivals}{rivals}  \mybox{3_warn}{warn}  \mybox{3_of}{of}  \mybox{3_possible}{possible}  \mybox{3_fraud}{fraud}  \mybox{3_semi}{;} \mybox{3_government}{government}  \mybox{3_says}{says}  \mybox{3_chavez}{chavez}  \mybox{3_apos}{'} \mybox{3_s}{s} \mybox{3_defeat}{defeat}  \mybox{3_could}{could}  \mybox{3_produce}{produce}  \mybox{3_turmoil}{turmoil}  \mybox{3_in}{in}  \mybox{3_world}{world}  \mybox{3_oil}{oil}  \mybox{3_market}{market}  \mybox{3_period}{.} \mybox{3_SEP}{[SEP]}\\
\end{tabular}
};
\end{adjustbox}
\caption{Word-level adversarial attack vs. iterative masking on AGNews sample on BERT classifier. Both adversarial attack and iterative masking perturb the prediction after manipulating a single token.} 
\label{fig:comparison_adv_mask}
\end{figure}

\section{Discussion and Conclusion}
\label{sec:fid-disc}

In analyzing the iterative-masking mechanism by which faithfulness is applied to neural text classifiers, we propose that this property is more accurately described as ``sensitivity to iterative masking".  Models that score high on these measures produce more significant perturbations in the presence of fewer masked tokens --- a property that depends on initialization-specific behaviour on samples that are increasingly outside the training data manifold as tokens are masked.  Further, sensitivity to iterative masking can be framed as akin to an adversarial attack in cases where the removal of a salient word should not change the true class of a sample, adding a strong task-specific element to its consideration as well.

Based on these findings, we caution against interpreting sensitivity to iterative masking as a positive indicator of ``interpretability" when comparing text classification models.  In the case of text classification, such measures may spuriously differentiate models based on their response to out-of-domain samples, and in some cases may favour models that are less adversarially robust.  When faced with the question of ``how do we compare the interpretability of neural text classifiers", we provide 3 main recommendations:

%Those seeking to improve the suitability of their models for use in real-world systems should focus on measures of adversarial robustness and fairness, which provide a more reliable basis for comparison between models.  %This, and the other considerations, clearly highlight that the practice of comparing models using faithfulness as a positive quality warrants scrutiny.

%If absolutely necessary, average the results from many training runs (including pre-training) with different initializations and measure statistical significance.}
\begin{enumerate}
\item{Avoid comparing different models using methods that assume each model will respond consistently on partially masked samples.  Such responses appear initialization-specific, and masked samples fall further outside of the manifold of model's training data as tokens are masked.}
\item{Take into account the task dataset on which the model is being evaluated.  For some tasks, removing tokens may not change the true class of a sample, and therefore is easily framed as a form of adversarial attack.  In these cases, robustness to removing tokens may be a desirable quality related to adversarial robustness.}
\item{Consider the trade-offs of a model relying heavily on a small number of salient input tokens to make decisions.  Making decisions based on the minimal number of tokens may be less important than other considerations (adversarial robustness or fairness, for example).}
\end{enumerate}

Overall our findings suggest that comparing neural text classifiers using masking-based faithfulness measures as a proxy for interpretability carries significant risks, even if models have the same or similar architectures.  Measurements of faithfulness based on iterative masking are dependent on model-specific behaviour, and partially masked samples are often well outside the data manifold of the original training data.  Model comparisons based on responses to iterative masking should be considered in nuanced terms and performed carefully.  %Considering faithfulness measures through the more literal lens of ``sensitivity to iterative masking" and referring to our provided recommendations may be helpful in this regard.

% taking into account the datasets used, and interpreted through the lens of robustness to iterative masking.

Based on our research, we have noted that successful text adversarial attacks result in highly salient features, and often result in increased fidelity scores on successful adversarial attacks.  This aligns with previous research that has used similar features for detecting adversarial attacks \citep{huber2022detecting}.  Beyond this observation, our findings indicate that adversarial training of neural text classifiers does not have a consistent impact on fidelity scores, suggesting that the relationship between adversarial training and robustness to iterative masking is less direct than previously thought.

%Using methods that reward  some of the The importance of whether a model relies on a single token if a text classification model can produce explanations that reflect the internal workings of the model, interpretability is not improved by having a model change prediction after masking 2 token vs 3 tokens.  Training a model on partially-masked samples to invert predictions could allow us to ``game" such measurements, without any real benefit.  More important than faithfulness, model creators should focus on improving adversarial robustness and fairness.  Faithfulness can be a useful tool for comparing explanations, but is often of limited utility for meaningfully comparing models.

%\subsection{Future Work}

%\subsubsection{Comparing Model Interpretability}
Significant future work exists in understanding neural text classifier interpretability.  Within neural text classification, there remain challenges in comparing faithfulness while avoiding pitfalls of initialization-specific behaviour.  Approaches that are axiomatic and well-principled, such as perturbing input features individually and measuring correlation between input feature rankings and changes in class confidence, themselves resemble explanation methods (e.g., integrated gradients \citep{sundararajan2017axiomatic}).  This similarity raises the concern of simply validating one explanation by using another.  Future evaluation work along these lines may focus on narrowing down the circumstances under which such comparisons may be reasonable.

%to what extent attempting to quantify the interpretability of a neural text classifier can result in simply validating one explanation method by using another, and under what circumstances this is reasonable.

%highlighting difficulties in disentangling data, model, and explanation method.

%When working with measures based on iterative masking, it is particularly important to keep this in mind.

%\subsubsection{Understanding Weaknesses in Neural Classifier Faithfulness}
% We could demonstrate the ridiculousness of this silly method of evaluating models by adversarially training a model to maximize classification inversions on the prescence of an "UNK" token.  If you masked by deleting tokens, you're at least classifying normal sentences, but it's even more obvious that you're just classifying a DIFFERENT SENTENCE ENTIRELY.  Especially if there's any inversions.  What a garbage amonut of rigour.

%Knowing whether a model changes prediction on a given sample after removing 1 token or after removing 3 tokens does not provide a meaningful assessment of interpretability.

%we argue that faithfulness is not a meaningful property for model comparison.  While faithfulness can be a useful property to measure when comparing neural text classifiers, all neural text classifiers on which it is possible to produce input feature attributions are effectively equally interpretable.  

%\FloatBarrier

%\bibliographystyle{splncs04}
\bibliography{abbr}

\end{document}